\documentclass{article}

    \usepackage[preprint]{neurips_2025}

\usepackage[utf8]{inputenc} 
\usepackage[T1]{fontenc}    
\usepackage{hyperref}       
\usepackage{url}            
\usepackage{booktabs}       
\usepackage{amsfonts}       
\usepackage{nicefrac}       
\usepackage{microtype}      
\usepackage{xcolor}         

\usepackage{graphicx}
\usepackage{mathtools}
\usepackage{algorithm}
\usepackage{algorithmic}

\title{Recovering Event Probabilities from Large Language Model Embeddings via Axiomatic Constraints}

\author{%
  Jian-Qiao Zhu \\
  Department of Computer Science\\
  Princeton University\\
  \texttt{jz5204@princeton.edu} \\
  \And
  Haijiang Yan \\
  Department of Psychology \\
  University of Warwick \\
  \AND
  Thomas L. Griffiths \\
  Departments of Psychology and Computer Science \\
  Princeton University \\
}

\begin{document}

\maketitle

\begin{abstract}
  Rational decision-making under uncertainty requires coherent degrees of belief in events. However, event probabilities generated by Large Language Models (LLMs) have been shown to exhibit incoherence, violating the axioms of probability theory. This raises the question of whether coherent event probabilities can be recovered from the embeddings used by the models. If so, those derived probabilities could be used as more accurate estimates in events involving uncertainty. To explore this question, we propose enforcing axiomatic constraints, such as the additive rule of probability theory, in the latent space learned by an extended variational autoencoder (VAE) applied to LLM embeddings. This approach enables event probabilities to naturally emerge in the latent space as the VAE learns to both reconstruct the original embeddings and predict the embeddings of semantically related events. We evaluate our method on complementary events (i.e., event A and its complement, event not-A), where the true probabilities of the two events must sum to 1. Experiment results on open-weight language models demonstrate that probabilities recovered from embeddings exhibit greater coherence than those directly reported by the corresponding models and align closely with the true probabilities.
\end{abstract}

\section{Introduction}

Understanding how language models represent and process information through their embeddings is a key focus in modern interpretability research \cite{elhage2021mathematical, gurnee2023finding, templeton2024scaling}. Of particular interest are features related to event probabilities, as these shape how models reason and make decisions when faced with uncertainty \cite{zheng2023judging, chen2018isolating, requeima2024llm, jia2024decision}. Emerging evidence suggests that event probabilities generated by LLMs in text responses (which we denote $\mathbf{P}_\text{judged}$) often lack coherence \cite{zhu2024incoherent}. For instance, when separately prompted about an event A and the complementary event not-A, the LLM-judged probabilities for A and not-A typically do not sum to 1 (i.e., $\mathbf{P}_\text{judged}(A) + \mathbf{P}_\text{judged}(\neg A) \neq 1$). In contrast, true probabilities inherently satisfy the axiomatic constraints dictated by the rules of probability theory, ensuring that they are always coherent. In addition to the rules of probability theory, LLMs have also been found to violate other axioms \cite{liu2024large, raman2024rationality, alsagheer2024comparing, horton2023large, binz2023using}.

Incoherence can have significant and potentially dangerous consequences. For instance, consider the risks of relying on an incoherent LLM for financial advice or allowing it to make investment decisions on one’s behalf. Incoherent beliefs, such as assigning probabilities like $P(\text{price up}) = P(\text{price down}) = 0.6$, can be easily exploited by arbitrageurs. Holding such incoherent beliefs may become ``money pumps,'' as inconsistent judgments create opportunities for others to profit at their expense. This phenomenon is explored in Dutch Book arguments, which emphasize the necessity of coherent beliefs to avoid such vulnerabilities \cite{pettigrew2020dutch}. Beyond finance, similarly adverse outcomes can arise when incoherent LLMs are applied to critical domains such as medicine and law.

While the probabilities produced in LLM responses may lack coherence, LLMs are generally capable of generating coherent text descriptions of complementary events \cite{brown2020language}. This suggests that the probabilities produced in LLM outputs may not directly correspond to the probabilities represented within their embeddings. Much as people might not always explicitly articulate their true thoughts, it is possible that LLMs do not generate probabilities that faithfully reflect their internal representations. In this paper, we investigate whether valid event probabilities can be recovered from model embeddings by imposing axiomatic constraints designed to enforce coherence.

Imposing axiomatic constraints on the latent space may initially appear challenging, as these constraints often involve complex relational dependencies among behaviors. However, we demonstrate that under mild assumptions, enforcing the additive rule for complementary events can be  achieved by enforcing structure in the latent space, akin to learning disentangled latent representations \cite{bengio2013deep, chen2018isolating, mathieu2019disentangling}.

Our contributions can be summarized as follows:
\begin{itemize}
    \item We propose a novel unsupervised learning method for extracting an interpretable latent space from LLM embeddings by enforcing axiomatic constraints.
    \item We show that a subset of latent variables, constrained by the axioms of probability theory, is associated with event probabilities encoded within LLM embeddings.
    \item The recovered probabilities demonstrate improved coherence and closely align with the true probabilities.
\end{itemize}

\begin{figure*}[t!]
    \centering
    \includegraphics[width=0.8\linewidth]{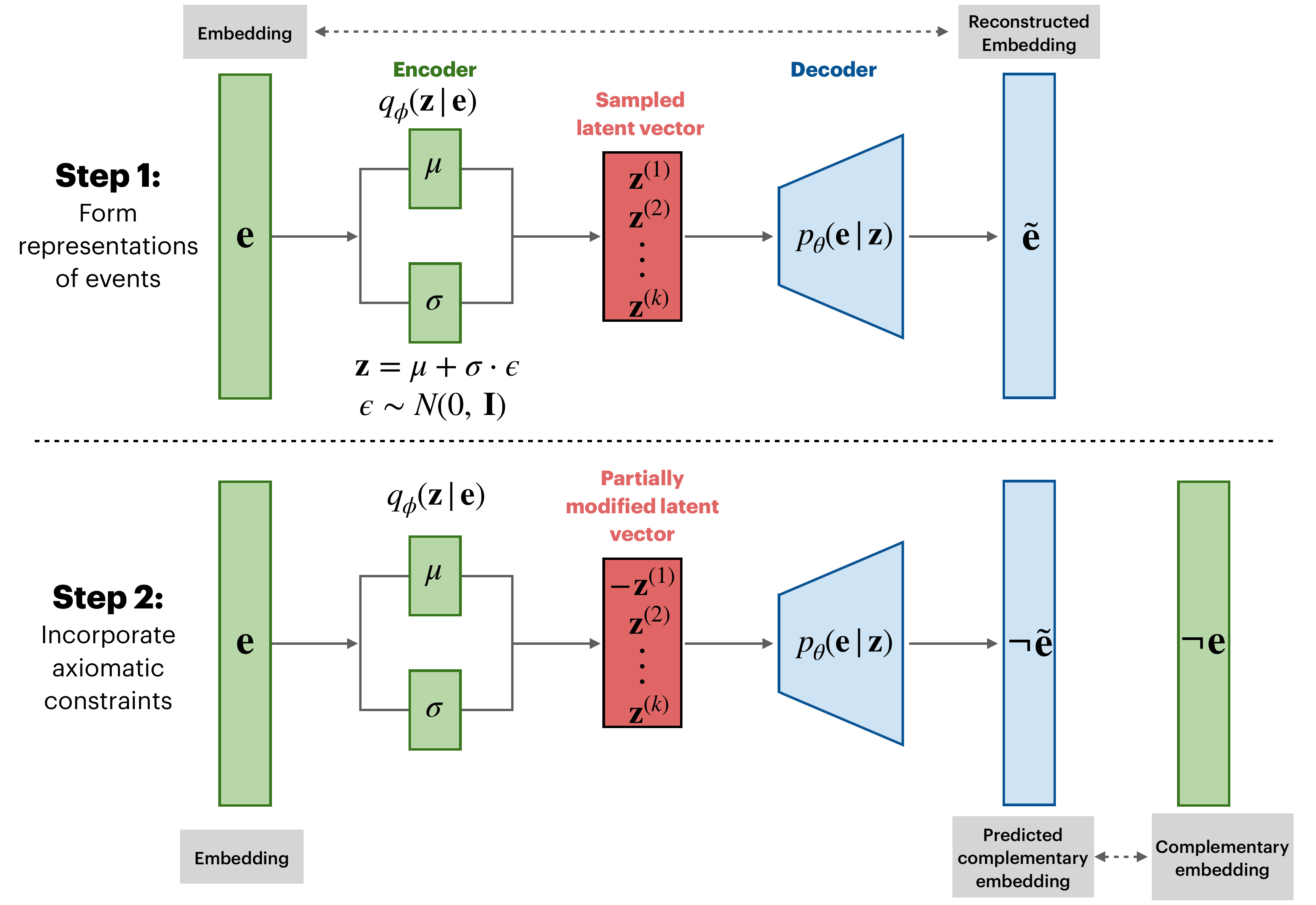}
    \caption{A schematic illustration of the two-step VAE-based learning algorithm. The objective is to distill axiomatic constraints between embeddings into the modified latent variables. 
    In the first step, a standard VAE is used to disentangle explanatory factors within the latent space from the model embeddings. This involves learning to reconstruct the original LLM embeddings, $\mathbf{e}$, by encoding them into a compressed latent space, $\mathbf{z}$, and subsequently decoding reconstructed embeddings, $\Tilde{\mathbf{e}}$. The latent space is modeled as a multivariate Gaussian distribution with mean $\mu$ and standard deviation $\sigma$. 
    In the second step, a subset of the latent variables is modified after encoding, and the modified latent vector is passed through the decoder to predict the embeddings of complementary events, $\neg \Tilde{\mathbf{e}}$.
    The probabilistic encoder and probabilistic decoder are denoted $q_\phi(\mathbf{z} |\mathbf{e})$ and $p_\theta(\mathbf{e} |\mathbf{z})$ respectively. Owing to the symmetry property (i.e., $\neg (\neg \mathbf{e}) = \mathbf{e}$), both $\mathbf{e}$ and $\neg \mathbf{e}$ can be used as input embeddings.}
    \label{fig:method-visual-2-step-AEVB}
\end{figure*}

\section{Problem Formulation}

Let us consider a dataset of $N$ prompts, each representing an event of interest, denoted as $\mathbf{S} = \{\mathbf{s}^{(i)}\}_{i=1}^N$. For example, a prompt might be $\mathbf{s}$: ``\textit{What is the probability of rolling a 5 on a 6-sided die?}'' Our goal is to investigate how LLMs represent event probabilities, relying solely on model embeddings and without access to the true probabilities of the events. Specifically, we aim to recover the latent representation of the event probability, $\mathbf{z}^{(i)}$, by extracting the corresponding embedding $\mathbf{e}^{(i)}$ from LLM, resulting in a dataset of embeddings: $\mathbf{E} = \{\mathbf{e}^{(i)}\}_{i=1}^N$.

Importantly, for each prompt, we also have a prompt for the complementary event: $\neg\mathbf{S} = \{\neg\mathbf{s}^{(i)}\}_{i=1}^N$. For the dice example above, the prompt for the complementary event would be $\neg\mathbf{s}$: ``\textit{What is the probability of rolling a number other than 5 on a 6-sided die?}'' Similarly, for this dataset of complementary events, we extract embeddings from the LLMs, forming a dataset of complementary event embeddings: $\neg\mathbf{E} = \{\neg\mathbf{e}^{(i)}\}_{i=1}^N$.

Within this formulation, we aim to address three interrelated problems:
\begin{itemize}
    \item Understanding how event probabilities are encoded within LLM embeddings.
    \item Developing efficient methods to recover and potentially enhance the representation of event probabilities in LLM embeddings.
    \item Exploring the implications of these learned representations when correct axiomatic constraints are imposed.
\end{itemize}

\section{Method}

Our unsupervised approach aims to learn a set of disentangled latent variables that serve two key purposes. First, these latent variables can be used to reconstruct the original embeddings. Second, through targeted modifications of specific latent variables, they can also be used to accurately generate the embeddings of complementary events. These modifications effectively encode the relationship between complementary events' embeddings--by activating or deactivating the modification, the same latent variables can predict either embedding. We can then enforce desired axiomatic constraints that respect the relationship between complementary events by imposing a specific prior structure on these modified latent variables. Conceptually, our approach aligns with methods that constrain neural networks to exploit the symmetry of problems through enforcing equivariance \cite{zaheer2017deep, cohen2016group}.

Specifically, given our focus on event probabilities from model embeddings $\mathbf{e} \in \mathbb{R}^d$, all other information within the embedding is considered extraneous and hence should be effectively compressed. To achieve this, we aim to map the original embedding into a smaller, more compact latent space $\mathbf{z} \in \mathbb{R}^k$, where $k \ll d$. Ideally, for simplicity and interpretability, the event probability would be encapsulated in a single latent variable (e.g., $\mathbf{z}^{(1)}$), with the remaining latent variables representing information unrelated to the event probability.

Using two datasets of model embeddings for complementary events, $\mathbf{E}$ and $\neg \mathbf{E}$, we employ the autoencoding variational Bayes algorithm to learn compressed latent representations $\mathbf{z}$. As illustrated in Figure \ref{fig:method-visual-2-step-AEVB} and Algorithm \ref{alg:two_step_VAE}, the proposed method consists of two interleaved steps during training. 

In the first step, we adopt the standard variational autoencoder (VAE) framework, where two neural networks separately implement the probabilistic encoder $q_\phi(\mathbf{z}|\mathbf{e})$ and decoder $p_\theta(\mathbf{e}|\mathbf{z})$ \cite{Kingma2014}. The parameters of these networks ($\phi$ and $\theta$) are jointly optimized to maximize the likelihood of the embeddings, $p(\mathbf{e})$. The prior distribution over latent variables is assumed to be a centered isotropic multivariate Gaussian, $p_\theta(\mathbf{z}) = \mathcal{N}(\mathbf{z}; 0, \mathbf{I})$.

In the second step, a key deviation from the standard VAE framework is introduced to capture the constraints that the axioms of probability theory impose on complementary events: this step optimizes the encoder and decoder parameters to maximize the probability of the complementary embeddings conditioned on the original embeddings, $p(\neg \mathbf{e}|\mathbf{e})$. The approach assumes that the constraint between $\neg \mathbf{e}$ and $\mathbf{e}$ can be captured by selectively modifying a subset of latent variables after encoding. Because the constraint is explicitly incorporated during the latent modifications, the encoder-decoder networks from Step 1 can be reused. Combined with the remaining unmodified latent variables, the partially modified latent vector is then passed through the decoder network to predict the embedding of the complementary event.

In summary, if the trained model successfully reconstructs the input embedding from a compressed latent representation and predicts the complementary embedding by partially modifying the same latent representation, the latent space should capture interpretable and meaningful low-dimensional information. Furthermore, the subset of latent variables that undergo modification should adhere to the constraints imposed during the modification process.

\begin{algorithm}[t!]
   \caption{Learning to recover event probabilities from LLM embeddings}
   \label{alg:two_step_VAE} 
\begin{algorithmic}[1] 
   \STATE $\phi, \theta \leftarrow$ Initialize parameters
   \REPEAT
    \STATE $\epsilon \leftarrow$ Random samples from standard Gaussian distribution
    \IF{Step 1 training}
    \STATE $\mathbf{E}^B \leftarrow$ Random minibatch of $B$ embeddings 
    \STATE $\mathbf{g} \leftarrow \nabla_{\phi, \theta} \mathcal{L}_1(\phi, \theta; \mathbf{E}^B, \epsilon)$ Gradients of minibatch estimator 
    \ENDIF
    \IF{Step 2 training}
    \STATE $\mathbf{E}^B, \neg\mathbf{E}^B \leftarrow$ Random minibatch of $B$ embeddings 
    \STATE $\mathbf{g} \leftarrow \nabla_{\phi, \theta} \mathcal{L}_2(\phi, \theta; \mathbf{E}^B,\neg\mathbf{E}^B, \epsilon, \phi_0)$ Gradients of minibatch estimator 
    \ENDIF
    \STATE $\phi, \theta \leftarrow$ Updating parameters using gradient $\mathbf{g}$
   \UNTIL{convergence of parameters ($\phi, \theta$)}
\end{algorithmic}
\end{algorithm}

\textbf{Variational bounds.}
Our learning algorithm aims to simultaneously capture two probability distributions: $p(\mathbf{e})$ and $p(\neg \mathbf{e} | \mathbf{e})$. We assume that model embeddings are generated through a random process involving latent variables $\mathbf{z}$. Approximating the intractable marginal likelihood $p(\mathbf{e})$ is achieved by jointly optimizing an encoder, $q_\phi(\mathbf{z}|\mathbf{e})$, and a decoder, $p_\theta(\mathbf{e}|\mathbf{z})$. Specifically, the optimization focuses on a variational lower bound of the marginal likelihood:
\begin{align}
    \log p(\mathbf{e}) & \geq \mathcal{L}_1(\phi, \theta; \mathbf{e})  = \mathbb{E}_{q_\phi(\mathbf{z}|\mathbf{e})} \Big[\log p_\theta(\mathbf{e}|\mathbf{z})\Big] - D_{KL}\Big( q_\phi(\mathbf{z}|\mathbf{e}) \parallel p_\theta(\mathbf{z}) \Big) \label{eq:objective_1}
\end{align}
The encoder and decoder can be parameterized using two neural networks (i.e., $\phi$ and $\theta$). The variational lower bound is differentiable with respect to these neural network parameters, enabling gradient-based optimization. To facilitate training, the reparameterization trick can be applied \cite{Kingma2014}.

Similarly, the variational bound for $p(\mathbf{e})$ can be extended to $p(\neg \mathbf{e} | \mathbf{e})$ (see Appendix \ref{ap:variational_bounds}):
\begin{align}
    \log p(\neg \mathbf{e} | \mathbf{e})
    & \geq \mathbb{E}_{q(\mathbf{z}|\mathbf{e})} \Big[\log p(\neg \mathbf{e}|\mathbf{z}, \mathbf{e})\Big] - D_{KL} \Big( q(\mathbf{z}|\mathbf{e}) \parallel p(\mathbf{z}|\mathbf{e}) \Big) \label{eq:objective_2}
\end{align}

This new bound introduces two additional quantities that are intractable. The generative distribution, $p(\neg \mathbf{e}|\mathbf{z}, \mathbf{e})$, can be simplified to depend solely on the latent variable. This assumes that the modifications in the latent space are sufficient to generate $\neg \mathbf{e}$:
\begin{align}
    p(\neg \mathbf{e}|\mathbf{z}, \mathbf{e}) & \approx p(\neg\mathbf{e} | T(\mathbf{z}))
\end{align}
where the modification operation consists of negating the first latent variable while preserving all others:  $T(\mathbf{z}) = [-\mathbf{z}^{(1)}, \mathbf{z}^{(-1)}]$.

Furthermore, the true posterior $p(\mathbf{z}|\mathbf{e})$ can be approximated using the posterior learned during the optimization of the variational bounds for $p(\mathbf{e})$:
\begin{align}
    p(\mathbf{z}|\mathbf{e}) & \approx q_\phi(\mathbf{z}|\mathbf{e})
\end{align}

Therefore, the training objective in the second step can be rewritten as:
\begin{align}
    \mathcal{L}_2(\phi, \theta; \mathbf{e}, \neg \mathbf{e}, \phi_0) & = \mathbb{E}_{q_\phi(\mathbf{z}|\mathbf{e})} \Big [ \log p_\theta(\neg\mathbf{e}|-\mathbf{z}^{(1)}, \mathbf{z}^{(-1)})\Big] - D_{KL} \Big( q_\phi(\mathbf{z}|\mathbf{e}) \parallel q_{\phi_0}(\mathbf{z}|\mathbf{e})) \Big) 
\end{align}
where $\phi_0$ denotes the parameters of the probabilistic encoder learned from the first step.

In short, we hypothesize that the relationship between embeddings for complementary events (i.e., the relationship between $\mathbf{e}$ and $\neg \mathbf{e}$) is captured in the first latent variable through the sign-flipping operation (i.e., the modification between $\mathbf{z}^{(1)}$ and $-\mathbf{z}^{(1)}$).  This implies that the probabilities of the two complementary events are intrinsically linked to the sign-flipping transformation of the first latent variable. Consequently, the first latent variable is best interpreted as representing log odds, rather than raw probabilities. Finally, the log odds recorded in the modified latent variable, $\mathbf{z}^{(1)}$, are converted back to the probability scale:
\begin{align}
    \mathbf{P}_\text{recovered} & = \frac{e^{\mathbf{z}^{(1)}}}{1+e^{\mathbf{z}^{(1)}}}  \label{eq:recover_prob}
\end{align}

By flipping the sign of $\mathbf{z}^{(1)}$, we enforce that the recovered probabilities satisfy the additive rule where the probabilities of complementary events must add up to 1:
\begin{align}
    \mathbf{P}_\text{recovered} + (1-\mathbf{P}_\text{recovered})  =  \frac{e^{\mathbf{z}^{(1)}}}{1+e^{\mathbf{z}^{(1)}}} + \frac{e^{-\mathbf{z}^{(1)}}}{1+e^{-\mathbf{z}^{(1)}}}
\end{align}

\section{Experiments}

Here, we demonstrate the effectiveness of our proposed method using a dataset of dice-related events. The primary advantage of using a probabilistic device like dice, as opposed to more ambiguous events, lies in the availability of their true probabilities. When the underlying true probabilities are known, the accuracy of probability estimates can be assessed by comparing them against these true values. This enables us to evaluate not only the coherence among a set of related probability estimates but also their alignment with the true probabilities.

We simulated all possible outcomes from rolling multi-sided dice, including scenarios such as a single roll of a 6-sided die, two rolls of a 6-sided die, and five rolls of a 4-sided die. In addition to determining the probability of observing a specific number or sum from the rolls, we also considered comparative queries, such as whether the sum is less than or greater than a given number (e.g., ``\textit{What is the probability that the sum of 8 rolls of a 10-sided die is greater than 15?}''). In total, we generated $N=1,728$ probability questions along with their corresponding complementary events.

Moreover, we constructed a test set consisting of 480 dice events. These events were generated using the same procedure as the training set but varied in the number of rolls and the number of sides on the die. Importantly, none of the 480 test events or their complements appeared in the training set. This test set constitutes approximately 28\% of the size of the training set (see Appendix \ref{ap:dataset_datails} for details).

We primarily focused on the Gemma-2-9b-instruct model, chosen for its open-weight architecture and robust performance on established benchmarks \cite{team2024gemma}. We replicated our experiments using Llama-3.1-8b-instruct \cite{dubey2024llama}, with detailed results presented in Appendix \ref{ap:llama}.

\begin{figure*}[t!]
    \centering
    \includegraphics[width=0.9\linewidth]{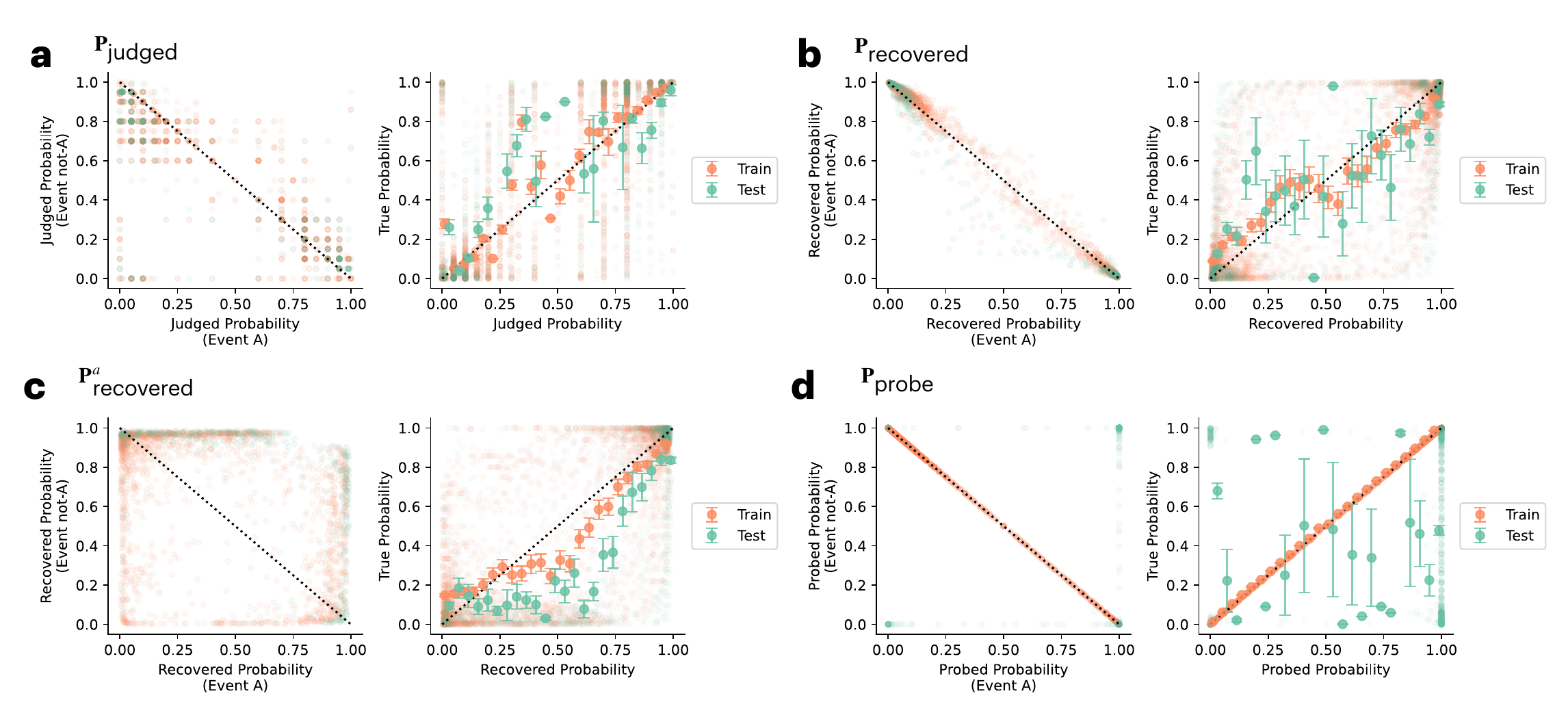}
    \caption{Comparison of event probability estimates elicited from Gemma-2-9b-instruct: (\textbf{a}) judged probabilities, (\textbf{b}) recovered probabilities, (\textbf{c}) recovered probabilities with Step 2 ablated, and (\textbf{d}) probabilities predicted by a linear probe. As indicated by the dotted black lines in the left panel, coherent probability estimates for event A and its complement (not-A) should sum to 1. The right panel compares the elicited probabilities with the true probabilities. Error bars represent standard errors of the mean for the window-binned scatter points.}
    \label{fig:main-results}
\end{figure*}

\textbf{Judged probabilities.}
We first evaluate the intuitive probability judgments generated in text by the Gemma model. To facilitate comparison with the probabilities recovered from model embeddings, we include the following system prompt before each probability question: ``\textit{You are asked to only provide an intuitive probability judgment as a number between 0 and 1.}'' The text responses were filtered to retain only a single real number between 0 and 1 representing the model’s judged probability. The model’s temperature was set to 1.

For complementary events, we define an incoherence measure by comparing the sum of the probabilities of the two events to the expected coherent sum of 1:
\begin{align}
    \text{Incoherence} = |\mathbf{P} + \neg\mathbf{P} - 1|
    \label{eq:incoherence}
\end{align}
This absolute difference quantifies the extent of incoherence in the probability estimates.

As shown in Table~\ref{tab:incoherence_results} and Figure~\ref{fig:main-results}a, $\mathbf{P}_\text{judged}$ exhibits mean incoherence scores of 0.13 on the training set and 0.14 on the test set, indicating a substantial deviation from the perfect coherence exhibited by $\mathbf{P}_\text{true}$. These findings replicate the results reported in \cite{zhu2024incoherent}. Despite this incoherence, $\mathbf{P}_\text{judged}$ demonstrates good alignment with the true probabilities, with Pearson's $r = 0.80~ (p < .01)$ and $r = 0.73 ~(p < .01)$ for the training and test sets, respectively.

To improve coherence, we also normalize $\mathbf{P}_\text{judged}$ by taking pairs of judgments for complementary events and dividing each by their summed total. This normalization substantially reduced incoherence while leaving accuracy, measured by MSE, relatively unchanged ($t(3455)=-2.54,~p=.01$ for the training set and $t(959)=-1.71, ~p=0.09$ for the test set). Furthermore, we explored an alternative approach in which complementary events were jointly presented within the same prompt, explicitly allowing the model to reason before providing its final probability judgments. This method significantly improved both the accuracy and coherence of probability judgments (see Appendix~\ref{ap:sentence_pairs}).

\begin{table*}[t!]
    \centering
    \caption{Quality of probability estimates (Gemma-2-9b-instruct). Incoherence is quantified using Equation~\ref{eq:incoherence}, while accuracy is assessed via Pearson’s correlation coefficient ($r$) and mean squared error (MSE) with respect to the true probabilities. Models that use LLM embeddings to predict event probabilities are trained exclusively on the training set, with model weights held fixed during evaluation on the test set. } \vspace{1mm}
    \begin{tabular}{llccc}
    \toprule
         Dataset & Elicitation method &  Incoherence $\downarrow$ (95\% CI) & \textcolor{black}{Pearson's $r$ $\uparrow$} & \textcolor{black}{MSE $\downarrow$}  \\ \hline
        \textcolor{black}{Train}  &  $\mathbf{P}_\text{judged}$ & 0.1297 ([0.1218, 0.1376]) & 0.8032 $(p<.01)$ & 0.0601 \\
         & \textcolor{black}{$\mathbf{P}_\text{judged}$ (normalized)} & \textcolor{black}{0.0023 ([0.0000, 0.0046])} & \textcolor{black}{0.7862 $(p<.01)$} & \textcolor{black}{0.0654} \\
        & $\mathbf{P}_\text{recovered}$  &  0.0227 ([0.0211, 0.0243]) & 0.8264 $(p<.01)$ & 0.0587   \\
          & $\mathbf{P}_\text{recovered}^a$ & 0.2948 ([0.2827, 0.3069]) & 0.7610 $(p<.01)$ & 0.0750   \\
         & \textcolor{black}{$\mathbf{P}_\text{probe}$} &  \textcolor{black}{0.0006 ([0.0006, 0.0007])} & \textcolor{black}{1.0 $(p<.01)$} & \textcolor{black}{$<0.0001$} \\ \hline
        \textcolor{black}{Test} & \textcolor{black}{$\mathbf{P}_\text{judged}$}  & \textcolor{black}{0.1366 ([0.1190, 0.1542])} & \textcolor{black}{0.7286 $(p<.01)$} & \textcolor{black}{\textbf{0.0927}}\\ 
         & \textcolor{black}{$\mathbf{P}_\text{judged}$ (normalized)} & \textcolor{black}{\textbf{0.0021} ([-0.0020, 0.0062])} & \textcolor{black}{0.7091$(p<.01)$} & \textcolor{black}{\textbf{0.1007}} \\
         & \textcolor{black}{$\mathbf{P}_\text{recovered}$}  & \textcolor{black}{0.0383 ([0.0314, 0.0452])} & \textcolor{black}{\textbf{0.7328} $(p<.01)$} & \textcolor{black}{\textbf{0.1014}}  \\ 
         & \textcolor{black}{$\mathbf{P}_\text{recovered}^a$} &  \textcolor{black}{0.3457 ([0.3223, 0.3692])} & \textcolor{black}{0.7167 $(p<.01)$} & \textcolor{black}{0.1140} \\ 
         & \textcolor{black}{$\mathbf{P}_\text{probe}$}  & \textcolor{black}{0.8535 ([0.8241, 0.8829])} & \textcolor{black}{-0.1328 $(p<.01)$} & \textcolor{black}{0.4654}\\
    \bottomrule
    \end{tabular}\\ \vspace{1mm}
    \textit{Note.} $\mathbf{P}_\text{recovered}^a$ refers to the recovered event probabilities from the ablated model. $\mathbf{P}_\text{probe}$ refers to the probabilities predicted by a linear probe. Bolded numbers indicate the best-performing methods on the test set. When multiple methods are highlighted, their performance differences are not statistically significant at the 0.05 level.
    \label{tab:incoherence_results}
\end{table*}

\begin{figure*}[t!]
    \centering
    \includegraphics[width=0.9\linewidth]{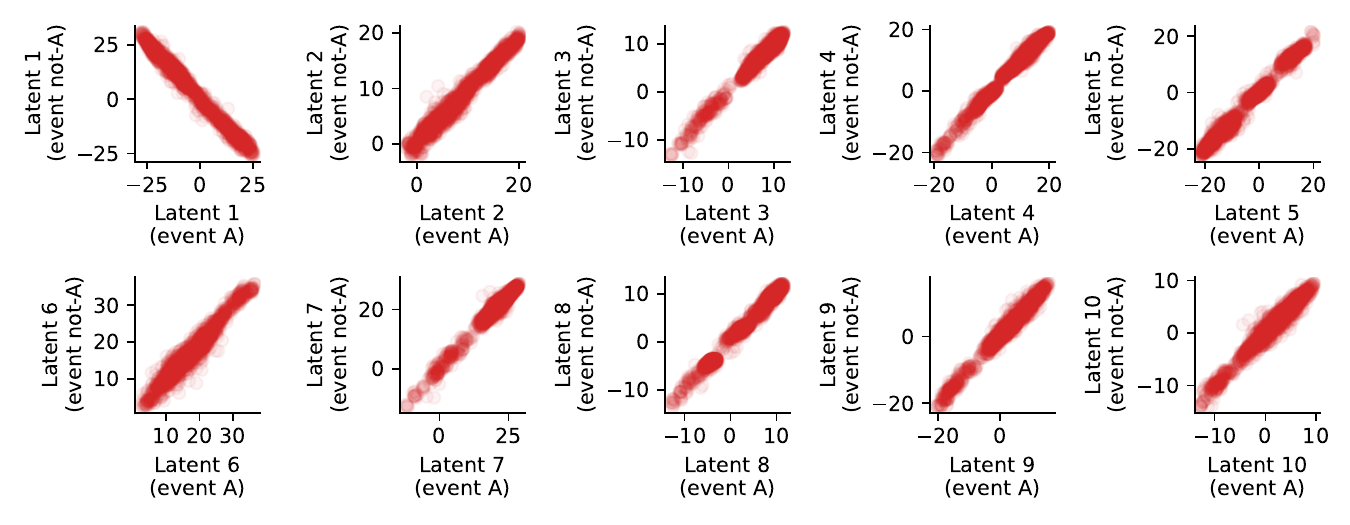}
    \caption{Visualization of the mean values of the 10 latent variables. Each panel compares the mean of the latent variables for embeddings corresponding to event A (horizontal axis) with those for embeddings corresponding to event not-A (vertical axis). Only the modified latent variable 1 exhibits a negative relationship. }
    \label{fig:latent-visual}
\end{figure*}

\textbf{Recovered probabilities.}
Now we apply unsupervised methods on model embeddings in order to learn to recover interpretable event probabilities. For each probability question, we extracted the embedding of the last token from the final layer before the logit computation, leading to two datasets of $\{\mathbf{e}^{(i)}\}_{i=1}^{1728}$ and $\{\neg\mathbf{e}^{(i)}\}_{i=1}^{1728}$. Notably, neither $\mathbf{P}_\text{true}$ nor $\mathbf{P}_\text{judged}$ is provided during this process. Due to the symmetry property (i.e., $\neg (\neg \mathbf{e}) = \mathbf{e}$), both $\mathbf{e}$ and $\neg\mathbf{e}$ were used as input embeddings.

Since the modified latent variable is interpreted as representing log odds, the centered isotropic Gaussian prior $p_\theta(\mathbf{z}) = \mathcal{N}(\mathbf{z}; 0, \mathbf{I})$ effectively enforces the axiomatic constraint of complementary events. This is because the additive rule for complementary events in the log-odds space corresponds to the sum of the log odds of the two events equaling zero. Therefore, the Gaussian prior with a mean of zero can be interpreted as a form of regularization that aligns the latent variables with the imposed axiomatic constraint. As a corollary, stricter regularization can be imposed by increasing the weight of the divergence from this prior, more strongly enforcing adherence to the axiomatic constraint.

In practice, we employ $\beta$-VAE \cite{higgins2017beta} to control the regularization strength toward the desired axiomatic constraint. $\beta$-VAE introduces a hyperparameter, $\beta$, which scales the KL divergence term in the VAE objective (see Equation \ref{eq:objective_1} and \ref{eq:objective_2}). In our experiments, we set $\beta=5$ to emphasize regularization toward the Gaussian prior. When converting log odds back to the probability scale, we applied the same value of 5 for temperature. 
Step 1 and Step 2 of the training process were interleaved every 10 training episodes. In each episode, a batch of 128 embeddings, along with those of the complementary events, was randomly selected. AdamW with a learning rate of 1e-4 was used for optimization \cite{loshchilov2017decoupled}. 

The architecture consists of an encoder and decoder, each implemented as a 3-layer feedforward neural network with GELU activation functions \cite{hendrycks2016gaussian}. Each layer contains $d$ neurons, where $d$ is the embedding dimension, and the hidden dimension is set to 10.

Figure \ref{fig:latent-visual} presents the mean values of the 10 latent variables after training, comparing the embeddings for the two complementary events. All latent variables exhibit a positive relationship, except for the modified latent variable (i.e., $\mathbf{z}^{(1)}$). This suggests that the relationship of complementary events between $\mathbf{e}$ and $\neg\mathbf{e}$ has been successfully distilled into this specific latent variable. As illustrated in Figure \ref{fig:main-results} (second row), the log odds values of $\mathbf{z}^{(1)}$ were transformed back into the probability scale, yielding $\mathbf{P}_\text{recovered}$ (see Equation \ref{eq:recover_prob}).

On the held-out test set (see Table \ref{tab:incoherence_results}), we find that $\mathbf{P}_\text{recovered}$ is significantly more coherent than $\mathbf{P}_\text{judged}$, as indicated by a lower incoherence score ($t(479) = -10.47, p < .01$). However, its coherence is statistically indistinguishable from that of the normalized $\mathbf{P}_\text{judged}$ ($t(479) = 1.01, p = .31$). In terms of accuracy, measured by MSE, $\mathbf{P}_\text{recovered}$ performs comparably to $\mathbf{P}_\text{judged}$ ($t(959) = -1.58, p = .12$). Additionally, $\mathbf{P}_\text{recovered}$ and $\mathbf{P}_\text{judged}$ are strongly correlated (Pearson’s $r = 0.90, p < .01$), indicating substantial alignment between the two (see Appendix \ref{ap:judge_vs_recovered}). Taken together, these results suggest that while LLM embeddings encode information that supports more coherent probability estimates, there remains considerable room for improvement in how event probabilities are represented.

\textbf{Non-modified latent variables.}
We examine the remaining nine non-modified latent variables, $\mathbf{z}^{(2:10)}$, which exhibited a positive relationship between $\mathbf{e}$ and $\neg\mathbf{e}$. To assess whether these latent variables are interpretable, we regress the features of the prompts onto the mean values of these latent variables. For each prompt, we identified six features, as detailed in Table \ref{tab:lasso_regression} features column. To illustrate, consider the prompt ``\textit{What is the probability of rolling a 5 on a 6-sided die?}''. This prompt can be characterized by the following six features: (i) one roll, (ii) a 6-sided die, (iii) an outcome of 5, (iv) a true probability of 1/6, (v) no reference to a sum, and (vi) a comparison type of ``equal to.''

Using Lasso regression with a penalty ratio of 1, we find that many non-modified latent variables are associated with specific features of the prompts. As shown in Table \ref{tab:lasso_regression} (see Appendix \ref{ap:lasso}), $\mathbf{z}^{(9)}$ is strongly associated with the number of rolls, accounting for 50\% variance ($R^2=50\%$); $\mathbf{z}^{(7)}$ corresponds to whether a prompt involves a summation of multiple rolls ($R^2=61\%$); and $\mathbf{z}^{(5)}$ is linked to the comparison type ($R^2=61\%$). In addition, true probabilities and outcomes appear to be influenced by a combination of latent variables, with the true probabilities being predominantly affected by the modified latent variable, $\mathbf{z}^{(1)}$. While the modified $\mathbf{z}^{(1)}$ is mostly correlated with the true probabilities, certain other latent variables, such as $\mathbf{z}^{(5,7,9)}$, exhibit correlations with multiple features. This suggests the presence of \textit{polysemanticity} \cite{bolukbasi2021interpretability, elhage2022toy}, a phenomenon where neuron activates for a range of different features, in the compressed latent space.

\textbf{Ablation study.}
To evaluate the effectiveness of our proposed two-step method, we conducted an ablation study. A key concern is whether Step 2 in Figure \ref{fig:method-visual-2-step-AEVB}, which modifies the latent vector for more targeted isolation in the latent space, is necessary to identify latent variables associated with event probabilities. Specifically, we sought to determine whether a disentangled representation could be achieved by applying a standard VAE to LLM embeddings without Step 2.

To address this question, we removed the interleaved training from our method and directly implemented a $\beta$-VAE using the same hyperparameters and architecture. In this ablated model, the objective is learning to reconstruct the input embeddings. Notably, both $\mathbf{e}$ and $\neg\mathbf{e}$ were used as input embeddings, ensuring that the ablated model was trained on the same amount of data as  before. 

The learned latent representations are shown in Figure \ref{fig:latent-visual-ablated} (see Appendix \ref{ap:visual_ablated_model}). These latent representations do not exhibit clear relationships (positive or negative) between $\mathbf{e}$ and $\neg \mathbf{e}$. This suggests that the phenomenon of polysemanticity is more pronounced in representations learned from the standard $\beta$-VAE compared to those produced by our two-step method. Consequently, it is challenging to identify latent variables in the ablated model that are specifically associated with event probabilities. While incorporating additional supervision signals from true probabilities could potentially enable the construction of a linear model that combines multiple latent variables to predict event probabilities, this would undermine the objective of recovering event probabilities through unsupervised learning.

Nonetheless, the latent variable associated with event probabilities is expected to exhibit the strongest negative correlation between complementary events. Consistent with this expectation, we found the 9th latent variable most negatively correlated (Pearson's $r=-0.31,~ p<.01$). The recovered probabilities from the ablated model (which we denote $\mathbf{P}_\text{recovered}^a$) were derived based on this latent variable (see Figure \ref{fig:main-results}c). However, $\mathbf{P}_\text{recovered}^a$ are less coherent than $\mathbf{P}_\text{recovered}$ ($t(1727)=-44.68,~ p<.01$ for incoherence scores) and also less accurate ($t(3455)=-6.71,~ p<.01$ for MSE) in the training set. These findings suggest that ablating Step 2 results in a less interpretable latent space.

\textbf{Linear Probe.}
Finally, we train a linear probe to map LLM embeddings onto the log-odds of the true probabilities. Specifically, $\text{logit}(\mathbf{P}_\text{true}) = \mathbf{\beta}^\top \mathbf{e}$, where $\mathbf{\beta}$ denotes the regression coefficients corresponding to the LLM embedding dimensions. At test time, these coefficients are applied to embeddings from held-out events, and the predicted logits are transformed back to probabilities. As shown in Figure~\ref{fig:main-results}d and Table \ref{tab:incoherence_results}, the probe achieves near-perfect coherence and accuracy on the training set but fails to generalize to the test set. In Appendix~\ref{ap:alternative_probe}, we examine an alternative approach using $\mathbf{P}_\text{true}$ directly as the training target. While this yields a more stable mapping from embeddings to probabilities, the resulting predictions on the test set are not guaranteed to be coherent, as they may fall outside the [0, 1] interval.

\section{Related Work}

\textbf{Mechanistic Interpretability.} 
Significant research efforts of modern interpretability research have focused on leveraging sparse autoencoders \cite{ng2011sparse} to uncover a neural network’s features by learning sparse decompositions from embeddings \cite{demircan2024sparse, cunningham2023sparse, wang2022interpretability}. This approach has successfully revealed interpretable features in model embeddings, such as representations of the Golden Gate Bridge, Python code errors, and human emotions extracted from Claude 3 Sonnet \cite{templeton2024scaling}. Although both sparse autoencoders and our method use unsupervised learning, our approach places greater emphasis on compressing the information in model embeddings into a significantly lower-dimensional space. As a result, our method provides more targeted and focused interpretations.

\textbf{Rationality and Axiomatic Constraints.} 
Rational behaviors are defined by adherence to a set of axioms. Kolmogorov’s probability theory formalizes probability measures through three fundamental axioms: (i) the probability of any event is non-negative, (ii) the probability of the entire sample space is 1, and (iii) if two events are disjoint, the probability of their union is the sum of their individual probabilities \cite{kolmogorov2018foundations}. A key implication of these axioms is the constraint that the probabilities of an event and its complement must sum to 1. 

While coherence provides a valuable measure of consistency across behaviors, machine learning research often prioritizes accuracy (or calibration) \cite{kumar2019verified, shrivastava2023llamas, guo2017calibration, ye2024benchmarking}, defined as the correspondence between predictions and true values. Coherence and accuracy in probability estimates are related, as true probabilities are simultaneously the most coherent and accurate \cite{pettigrew2016accuracy, zhu2022clarifying}. However, it is important to note that greater coherence does not necessarily imply higher accuracy. For instance, one might produce coherent probability judgments for a fair coin, such as  $p(\text{Head}) = 1 - p(\text{Tail}) = 0.1$ , which are consistent yet inaccurate, as the true probabilities for both Head and Tail are 0.5.

\textbf{Disentanglement.}
Our work is also closely related to research on disentanglement, which aims to uncover independent factors that explain the variability in data \cite{bengio2013deep, hyvarinen2023nonlinear}. In our approach, we extend this goal to explaining model embeddings. Disentangled representation learning typically involves imposing a prior structure on the learned representation \cite{chen2018isolating, kim2018disentangling, mathieu2019disentangling, burgess2018understanding}. In our method, this prior structure is directly tied to the imposed axiomatic constraints, where a centered Gaussian prior corresponds to the additive rule of probability theory. In this sense, our work not only extends disentanglement to the interpretation of model embeddings but also generalizes the prior structure of learned representations to enforce desirable axiomatic constraints.

\section{Discussion}

Our unsupervised approach recovers more coherent event probabilities from LLM embeddings than those obtained through prompting probability judgments in text. This improvement is achieved by enforcing the additive rule of probability theory on the embeddings of complementary events. Moreover, the resulting latent space exhibits high interpretability, with individual latent variables corresponding to specific features of the prompt. This suggests that when properly trained, the compressed latent representation derived from LLM embeddings can also effectively disentangle LLM embeddings without introducing sparsity bias on the latent space. 

\subsection{Limitations and Future Directions}

We note several limitations of our current work that suggest promising directions for future research. 


\textbf{Imposing more sophisticated axiomatic constraints.}
The latent space can be extended to accommodate additional axiomatic constraints. For instance, the relationship between conjunctive and constituent events could be enforced through Bayes' rule: $P(A \text{~and~} B) = P(A)P(B|A)$. Future work could explore methods for effectively implementing such axiomatic constraints.

\textbf{Causal interventions.}
A natural next step is to investigate whether editing neuron activations based on the learned latent variables can produce more coherent probability judgments. We consider this a promising direction for establishing a clearer causal relationship.

\textbf{Generalizing to other constraints and modalities.}
Our approach could be extended to explore various other constraints across different LLM modalities. For instance, in multimodal LLMs, embeddings should exhibit rotational invariance when processing images of the same object from different angles. Future research could investigate how to impose such constraints.

\subsubsection{Conclusion}

We have shown that coherent probability judgments can be extracted from LLMs by imposing axiomatic constraints on the underlying representations. The successful recovery of coherent probability estimates from LLM embeddings suggests that critical information might be lost when LLMs generate probability judgments in text form. Our findings support recent hypotheses that certain LLM capabilities -- in this case, the ability to generate more accurate probability judgments -- remain ``hidden'' when only examining model responses but can be unlocked through self-improvement techniques such as bootstrapping \cite{zelikman2022star}. It is also worth noting that the capabilities of LLMs for generating coherent judgments correlate with the quality of the recovered probabilities, as embeddings from more coherent LLM yield better recovered probabilities. This suggests that directly incorporating such constraints into LLMs might enhance their ability to reason effectively about probabilities.

\textbf{Acknowledgments.}
This work and related results were made possible with the support of the NOMIS Foundation. H. Yan acknowledges the Chancellor's International Scholarship from the University of Warwick for additional support.

\bibliography{references}
\bibliographystyle{plain}








\appendix

\section{Variational bounds}
\label{ap:variational_bounds}

Here we present a detailed derivation of the variational bounds reported in the main text. First, we review the bound established for variational Bayes \cite{Kingma2014}, adapting the notation to our context where the input data are LLM embeddings (i.e., $\mathbf{e}$):
\begin{align}
    \log p(\mathbf{e}) & = \log \int p(\mathbf{e},\mathbf{z}) d\mathbf{z} \\
    & = \log \int q(\mathbf{z}|\mathbf{e})\frac{p(\mathbf{e},\mathbf{z})}{q(\mathbf{z}|\mathbf{e})} d\mathbf{z} \\
    & \geq \int q(\mathbf{z}|\mathbf{e}) \log \frac{p(\mathbf{e},\mathbf{z})}{q(\mathbf{z}|\mathbf{e})} d\mathbf{z} \\
    & = \int q(\mathbf{z}|\mathbf{e}) \log p(\mathbf{e}|\mathbf{z})d\mathbf{z} + \int q(\mathbf{z}|\mathbf{e})\log \frac{p(\mathbf{z})}{q(\mathbf{z}|\mathbf{e})} d\mathbf{z} \\
    & = \mathbb{E}_{q(\mathbf{z}|\mathbf{e})} \Big[\log p(\mathbf{e}|\mathbf{z})\Big] - D_{KL}\Big( q(\mathbf{z}|\mathbf{e}) \parallel p(\mathbf{z}) \Big)
\end{align}

We can generalize to the problem of predicting complementary embeddings, yielding the following variational bound:
\begin{align}
    \log p(\neg \mathbf{e} | \mathbf{e}) & = \log \int p(\neg \mathbf{e},\mathbf{z} | \mathbf{e}) d\mathbf{z} \\
    & = \log \int q(\mathbf{z}|\mathbf{e})\frac{p(\neg \mathbf{e},\mathbf{z}|\mathbf{e})}{q(\mathbf{z}|\mathbf{e})} d\mathbf{z} \\
    & \geq \int q(\mathbf{z}|\mathbf{e}) \log \frac{p(\neg \mathbf{e},\mathbf{z}|\mathbf{e})}{q(\mathbf{z}|\mathbf{e})} d\mathbf{z} \\
    & = \int q(\mathbf{z}|\mathbf{e}) \log p(\neg \mathbf{e}|\mathbf{z}, \mathbf{e})d\mathbf{z} + \int q(\mathbf{z}|\mathbf{e})\log \frac{p(\mathbf{z}|\mathbf{e})}{q(\mathbf{z}|\mathbf{e})} d\mathbf{z} \\
    & = \mathbb{E}_{q(\mathbf{z}|\mathbf{e})} \Big[\log p(\neg \mathbf{e}|\mathbf{z}, \mathbf{e})\Big] - D_{KL} \Big( q(\mathbf{z}|\mathbf{e}) \parallel p(\mathbf{z}|\mathbf{e}) \Big)
\end{align}

\section{Additional details on datasets}
\label{ap:dataset_datails}

We provide here the detailed procedure used to construct the training and test sets. Dice-related events are denoted using the notation $[x]$d$[y]$, which refers to rolling $x$ dice, each with $y$ sides. For the training set, we included the following event types: 1d6, 2d6, 3d6, 4d6, 5d6, 1d5, 2d5, 3d5, 4d5, 5d5, 1d7, 2d7, 3d7, 4d7, 5d7, 1d3, 2d3, 3d3, 4d3, 5d3, 1d4, 2d4, 3d4, 4d4, 5d4, 1d10, 2d10, 3d10, 1d17, and 2d17. This yielded a total of 1,728 unique probability questions, each paired with its complementary event. For the test set, we selected the following held-out event types: 6d6, 4d10, 3d12, and 2d16, resulting in 480 probability questions, each also paired with its complement.

\section{Probability judgments with complementary events in context}
\label{ap:sentence_pairs}
In this exploratory analysis, we investigate whether prompting the Gemma model with complementary event pairs together and allowing it to explicitly reason before producing probability estimates leads to improved judgments. Specifically, we present both an event (A) and its complement (not-A) simultaneously within a single prompt, instructing the model to output two probability judgments in JSON format to facilitate extraction. Moreover, the model is permitted to provide intermediate reasoning steps before finalizing its responses in JSON.

Evaluating this approach on the dice events from our test set, we find significant improvements in both coherence and accuracy of judged probabilities. The mean incoherence score is effectively zero, indicating that the model rarely produces complementary probability judgments that fail to sum to one. Furthermore, judged probabilities elicited in this context achieve a Pearson correlation of $r = 0.81 ~(p < .01)$ and a MSE of 0.07 relative to the true probabilities. This MSE is significantly lower than that observed when probabilities are elicited separately for each event ($t(959) = -5.11,~ p < .01$), suggesting that the model generates more accurate probability estimates when complementary events are presented together and reasoning is explicitly encouraged.

\section{Relationship between judged and recovered probabilities}
\label{ap:judge_vs_recovered}

\begin{figure}[h!]
    \centering
    \includegraphics[width=0.5\linewidth]{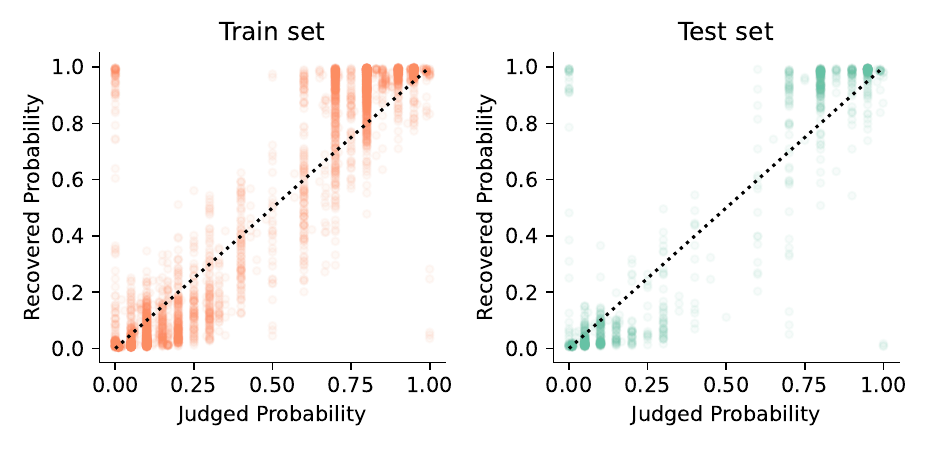}
    \caption{Relationship between judged probabilities (horizontal axis) and recovered probabilities (vertical axis). The dashed black lines indicate perfect agreement. 
    }
    \label{fig:judge_vs_recovered}
\end{figure}

To complement our main analysis, we examine the relationship between judged and recovered probabilities. As illustrated in Figure~\ref{fig:judge_vs_recovered}, these two groups of probabilities exhibit strong correspondence.

\section{Latent space is interpretable}
\label{ap:lasso}

In this section, we conduct a series of Lasso regressions using the latent variables learned from our model. Table \ref{tab:lasso_regression} summarizes the results.

\begin{table*}[h!]
    \centering
    \caption{Lasso regression of the latent variables onto interpretable features of the prompts involving dice-related events.} \vspace{1mm}
    \setlength{\tabcolsep}{3pt}
    \begin{tabular}{lccccccccccc}
    \toprule
        Features & $\mathbf{z}^{(1)}$  & $\mathbf{z}^{(2)}$  & $\mathbf{z}^{(3)}$  & $\mathbf{z}^{(4)}$  & $\mathbf{z}^{(5)}$  & $\mathbf{z}^{(6)}$  & $\mathbf{z}^{(7)}$  & $\mathbf{z}^{(8)}$ & $\mathbf{z}^{(9)}$ & $\mathbf{z}^{(10)}$ & $R^2$ \\ \hline
        \# of rolls & . & . & . & . & . & . & . & . & 0.10 & . &  0.50 \\
        \# of sides & . & . & . & . & . & . & -0.09 & . & . & . & 0.06\\
        Outcomes & . & . & 1.01 & -1.61 & -0.76 & . & . & . & 0.84 & . & 0.68 \\
        True probabilities & 2.08 & -0.08 & . & . & . & -0.10 & . & 0.04 & . & . & 0.68\\
        Sum$^a$  & . & . & . & . & . & . & 0.02 & . & . & . & 0.61 \\
        Comparison$^b$  & . & . & . & . & -0.05 & . & . & . & . & . & 0.61\\
    \bottomrule
    \end{tabular}\\ \vspace{1mm}
    \textit{Note.} Each row is an independent lasso regression with the penalty ratio set to 1. $^a$ This binary feature specifies whether the event pertains to the sum of multiple rolls or the outcome of a single roll. $^b$ This feature represents three types of comparisons: ``$<$ or $\leq$'', ``='', and ``$>$ or $\geq$''.
    \label{tab:lasso_regression}
\end{table*}

\section{LLM embeddings from other intermediate layers}

We further investigated intermediate layers of the Gemma-2-9b-instruct model by examining embeddings of the final token sampled at various layers (see Table \ref{tab:other_layers}). Moving from the last layer (layer 41) to earlier layers, we observed that the cosine similarity between embeddings of complementary event pairs generally decreased, indicating reduced representational similarity in earlier layers. Incoherence scores for $\mathbf{P}_\text{recovered}$ remained roughly consistent across layers. However, Pearson correlations measuring the correspondence between $\mathbf{P}_\text{recovered}$ and both $\mathbf{P}_\text{true}$ and $\mathbf{P}_\text{judged}$ tended to decrease at earlier layers. Notably, from layer 25 onwards (towards earlier layers), these correlation coefficients became negative.

\begin{table}[h!]
    \centering
    \caption{Probabilities recovered from embeddings at different layers of Gemma-2-9b-instruct. Cosine similarity values represent the mean and standard deviation computed between embeddings of complementary event pairs. Incoherence is quantified using Equation~\ref{eq:incoherence} based on the recovered probabilities. Pearson correlations ($r$) were calculated between recovered and judged probabilities, as well as between recovered and true probabilities. All metrics reported here are based exclusively on the training set. }\vspace{1mm}
    \setlength{\tabcolsep}{2pt}
    \begin{tabular}{lcccccc}
    \toprule
        Layer & Cosine similarity(SD) & Incoherence [95\%CI]$\downarrow$ & $r(\mathbf{P}_\text{recovered}, \mathbf{P}_\text{judged})\uparrow$ & $r(\mathbf{P}_\text{recovered}, \mathbf{P}_\text{true})\uparrow$ &  \\ \hline
        41 & 0.9731 (0.0101) & 0.0227 [0.0211, 0.0243] & 0.9099 & 0.8264 \\
        40 & 0.9450 (0.0214) & 0.0214 [0.0196, 0.0232] & 0.9051 & 0.8171 \\
        37 & 0.8428 (0.0597) & 0.0236 [0.0218, 0.0255] & 0.7979 & 0.8921 \\
        33 & 0.8278 (0.0678) & 0.0382 [0.0363, 0.0400] & 0.8810 & 0.7743 \\
        29 & 0.8457 (0.0533) & 0.0212 [0.0199, 0.0225] & 0.8580 & 0.7739 \\
        25 & 0.8704 (0.0370) & 0.0343 [0.0325, 0.0362] & -0.8683 & -0.7571 \\
        19 & 0.8491 (0.0654) & 0.0203 [0.0189, 0.0216] & -0.8745 & -0.7852 \\
        15 & 0.8750 (0.0491) & 0.0183 [0.0176, 0.0189] & -0.6752 & -0.5132 \\
        10 & 0.9372 (0.0166) & 0.0143 [0.0137, 0.0148]  & -0.4484 & -0.4672\\
    \bottomrule
    \end{tabular}
    \label{tab:other_layers}
\end{table}

\section{Latent space learned from the ablated model}

\label{ap:visual_ablated_model}
\begin{figure*}[h!]
    \centering
    \includegraphics[width=0.8\linewidth]{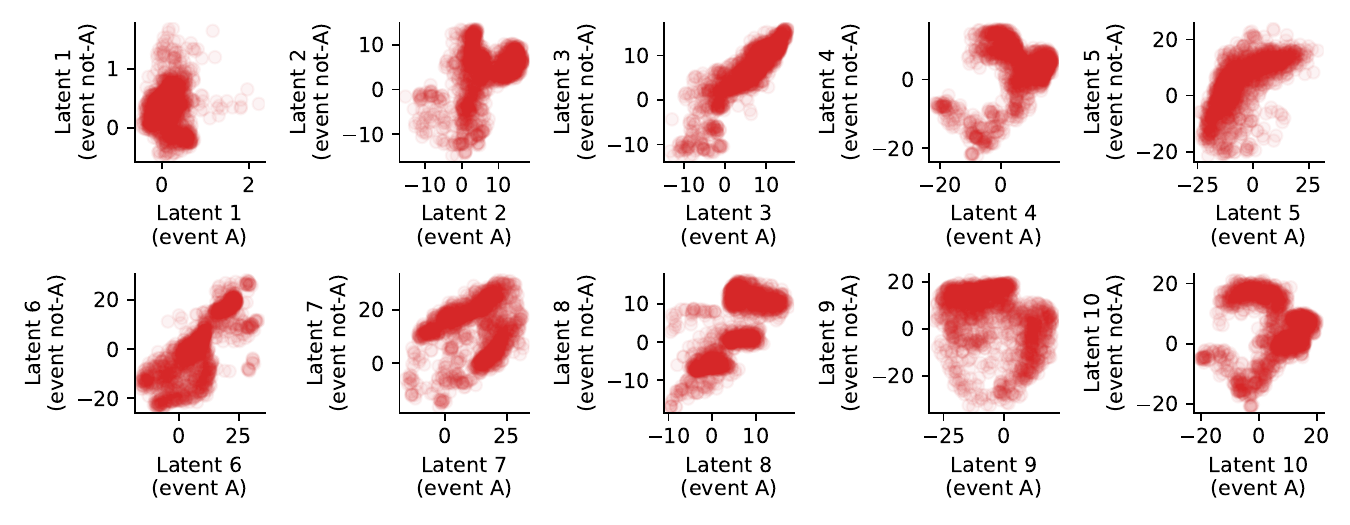}
    \caption{\textbf{Step 2 ablated.} Visualization of the mean values of the 10 latent variables. Each panel compares the mean of the latent variables for embeddings corresponding to event A (horizontal axis) with those for embeddings corresponding to event not-A (vertical axis). }
    \label{fig:latent-visual-ablated}
\end{figure*}

\section{Alternative Linear Probe}
\label{ap:alternative_probe}

\begin{figure}[h!]
    \centering
    \includegraphics[width=0.5\linewidth]{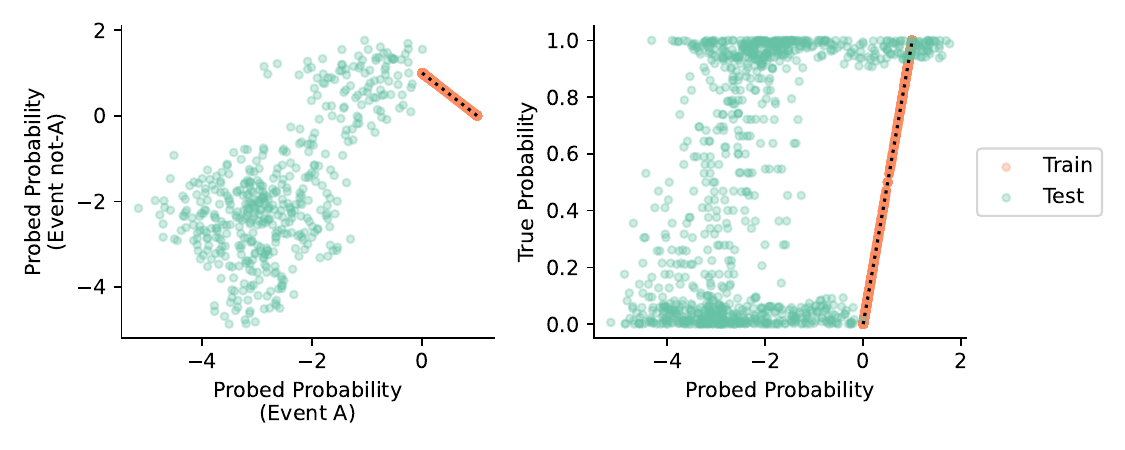}
    \caption{Alternative linear probe mapping LLM embeddings to true event probabilities. The left panel shows the relationship between the probed probabilities for complementary events, while the right panel compares the probed probabilities to the true probabilities. Black dashed lines indicate true event probabilities.}
    \label{fig:alternative_probe}
\end{figure}

In the main text, we examined a linear probe that maps LLM embeddings to the log-odds of the true event probabilities, following the model $\text{logit}(\mathbf{P}_\text{true}) = \beta^\top \mathbf{e}$. While effective on the training set, this approach yields unstable probability estimates at test time. Here, we consider an alternative probe that maps LLM embeddings directly to the true probabilities: $\mathbf{P}_\text{true} = \beta^\top \mathbf{e}$. Since this formulation outputs predictions on the probability scale, no further transformation is applied -- the probe’s output is interpreted directly as the predicted probability. As shown in Figure~\ref{fig:alternative_probe}, this alternative probe also achieves near-perfect performance on the training set but similarly fails to generalize to the test set.

\vspace{2cm}

\section{Recovering event probabilities from Llama-3.1-8b-instruct}
\label{ap:llama}

We conducted an additional experiment on Llama-3.1-8b-instruct \cite{dubey2024llama} using the same training set of dice-related events. We find that the Llama model performs significantly worse than Gemma-2-9b-instruct in judging dice-related events, both in terms of coherence and accuracy. The mean incoherence of $\mathbf{P}_\text{judged}$ from the Llama model is 0.24 (95\% CI [0.23, 0.25]), approximately twice as high as the incoherence observed in the Gemma model ($t(1727)=17.30, ~p<.01$). As shown in Table \ref{tab:accuracy_results_llama}, the correlation coefficient and MSE of $\mathbf{P}_\text{judged}$ from the Llama model compared to $\mathbf{P}_\text{true}$ are 0.67 and 0.10, respectively. $\mathbf{P}_\text{judged}$ from the Llamma model shows a higher MSE with respect to the true probabilities compared to the Gemma model ($t(3455)=11.09,~ p<.01$). Both measures indicate a poorer correspondence with the true probabilities compared to the judged probabilities generated by the Gemma model.


\begin{table}[H]
    \centering
    \caption{Comparisons between different probability estimates (Llama-3.1-8b-instruct).}\vspace{1mm}
    \begin{tabular}{lcc}
    \toprule
         & Pearson's $r$ $\uparrow$ & MSE $\downarrow$ \\ \hline
        $\mathbf{P}_\text{true}$ vs. $\mathbf{P}_\text{judged}$ & 0.6672 & 0.0966 \\
        $\mathbf{P}_\text{true}$ vs. $\mathbf{P}_\text{recovered}$ & 0.6527 & 0.1129  \\
        $\mathbf{P}_\text{judged}$ vs. $\mathbf{P}_\text{recovered}$ & 0.5624 & 0.1184  \\
    \bottomrule
    \end{tabular}\\ \vspace{1mm}
    \textit{Note.} MSE denotes mean square errors.
    \label{tab:accuracy_results_llama}
\end{table}

We applied our approach to the embeddings of Llama-3.1-8b-instruct to impose axiomatic constraints and recover event probabilities. Consistent with the main text, the embeddings were extracted from the last token in the final layer (layer 31) of the Llama model. The recovered event probabilities $\mathbf{P}_\text{recovered}$ showed improved coherence over $\mathbf{P}_\text{judged}$ ($t(1727)=29.91, p<.01$), with a mean incoherence of 0.07 (95\%CI [0.07, 0.08]). However, as shown in Table \ref{tab:accuracy_results_llama}, the accuracy of $\mathbf{P}_\text{recovered}$ measured in MSE was not improved compared to $\mathbf{P}_\text{judged}$ ($t(3455)=-3.94, p<.01$). 

Moreover, both the MSE and incoherence of $\mathbf{P}_\text{recovered}$ from the Llama model are significantly higher than those from the Gemma model (MSE: $t(3455)=16.14, p<.01$; incoherence: $t(1727)=23.40, p<.01$), indicating the recovered probability estimates from the Llama model are both less coherent and less accurate compared to those recovered from the Gemma model. These findings suggest that while coherence is generally enhanced, the LLM’s ability to generate high-quality estimates of event probabilities is crucial for enhancing the accuracy of the recovered probabilities using our approach.

\end{document}